\documentclass[runningheads]{llncs}
\usepackage{graphicx}
\usepackage{color}
\usepackage{subfigure}
\usepackage{multirow}
\usepackage[width=122mm,left=12mm,paperwidth=146mm,height=193mm,top=12mm,paperheight=217mm]{geometry}
\begin{document}
\pagestyle{headings}
\mainmatter
\def\ECCV18SubNumber{****}  

\title{A Segmentation-aware Deep Fusion Network for Compressed Sensing MRI} 

\titlerunning{A Segmentation-aware Deep Fusion Network for Compressed Sensing MRI}
\authorrunning{Fan et al}

\author{Zhiwen Fan$^{\ast}$, Liyan Sun$^{\ast}$, Xinghao Ding$^{\star}$, Yue Huang,\\ Congbo Cai, John Paisley$^{\dagger}$\\}

\institute{Fujian   Key   Laboratory   of   Sensing   and   Computing   for   Smart   City,  Xiamen   University, Fujian, China\\
$^{\dagger}$Department of Electrical Engineering, Columbia University, New York, NY, USA\\
$^{\star}$Correponding Author: dxh@xmu.edu.cn
\thanks{The co-first authors contributed equally.}}

\maketitle

\begin{abstract}
Compressed sensing MRI is a classic inverse problem in the field of computational imaging, accelerating the MR imaging by measuring less k-space data. The deep neural network models provide the stronger representation ability and faster reconstruction compared with "shallow" optimization-based methods. However, in the existing deep-based CS-MRI models, the high-level semantic supervision information from massive segmentation-labels in MRI dataset is overlooked. In this paper, we proposed a segmentation-aware deep fusion network called SADFN for compressed sensing MRI. The multilayer feature aggregation (MLFA) method is introduced here to fuse all the features from different layers in the segmentation network. Then, the aggregated feature maps containing semantic information are provided to each layer in the reconstruction network with a feature fusion strategy. This guarantees the reconstruction network is aware of the different regions in the image it reconstructs, simplifying the function mapping. We prove the utility of the cross-layer and cross-task information fusion strategy by comparative study. Extensive experiments on brain segmentation benchmark MRBrainS validated that the proposed SADFN model achieves state-of-the-art accuracy in compressed sensing MRI. This paper provides a novel approach to guide the low-level visual task using the information from mid- or high-level task.
\end{abstract}

\section{Introduction}
\label{Intro}

\begin{figure}[htb!]
\begin{center}
   \subfigure[\scriptsize Full-sampled]  {\label{fig1a} \includegraphics[width=0.193\textwidth]{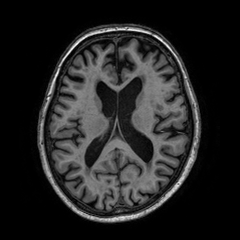}}
   \subfigure[\scriptsize Under-sampled] {\label{fig1b} \includegraphics[width=0.193\textwidth]{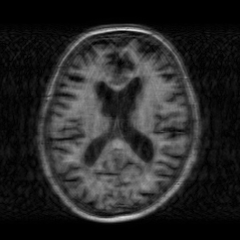}}
   \subfigure[\scriptsize Seg Label]   {\label{fig1c} \includegraphics[width=0.193\textwidth]{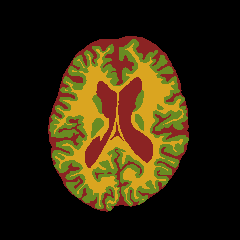}}\\
   \subfigure[\scriptsize Whole] {\label{fig1d} \includegraphics[width=0.18\textwidth]{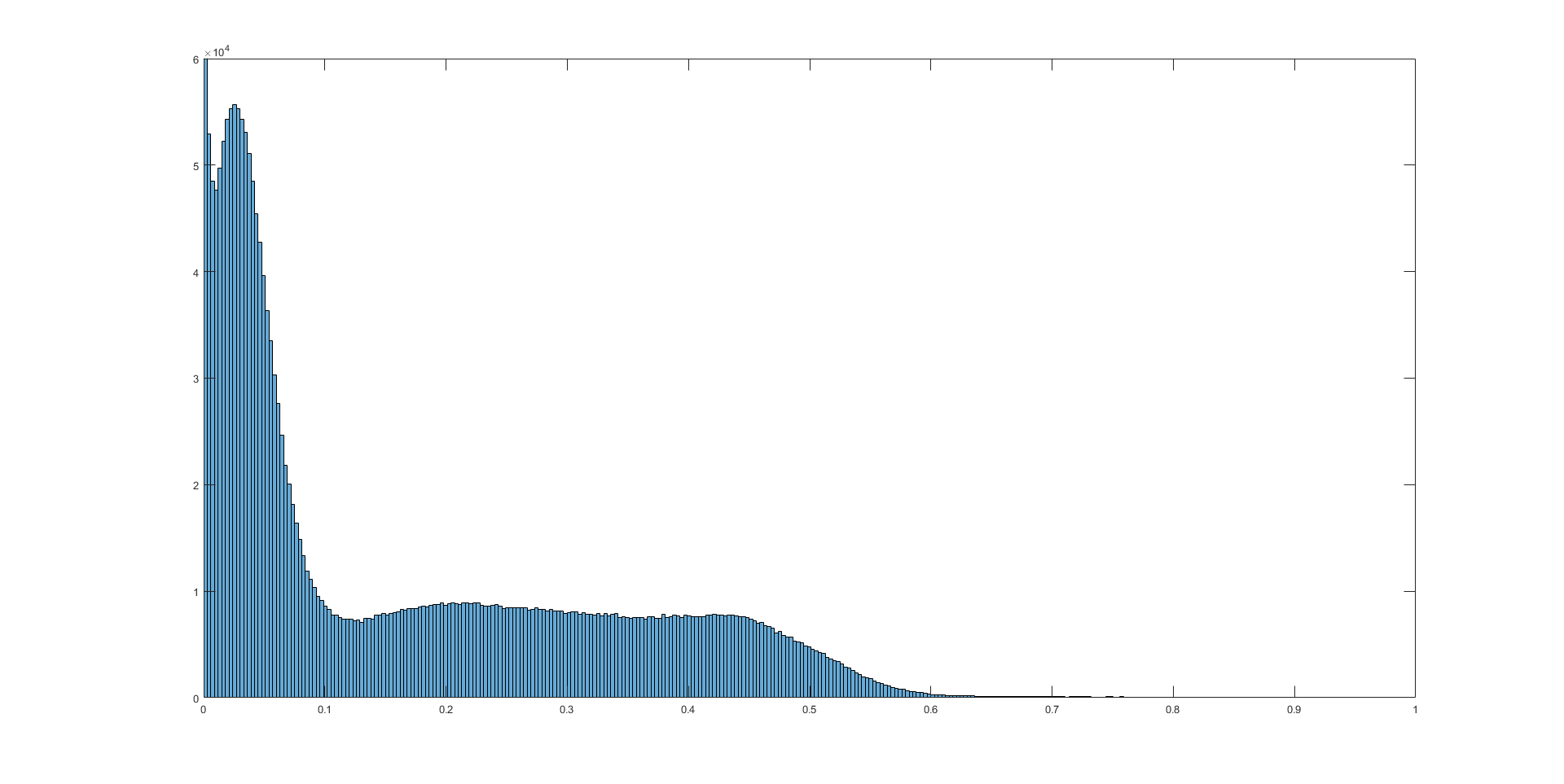}}
   \subfigure[\scriptsize BG]    {\label{fig1e} \includegraphics[width=0.18\textwidth]{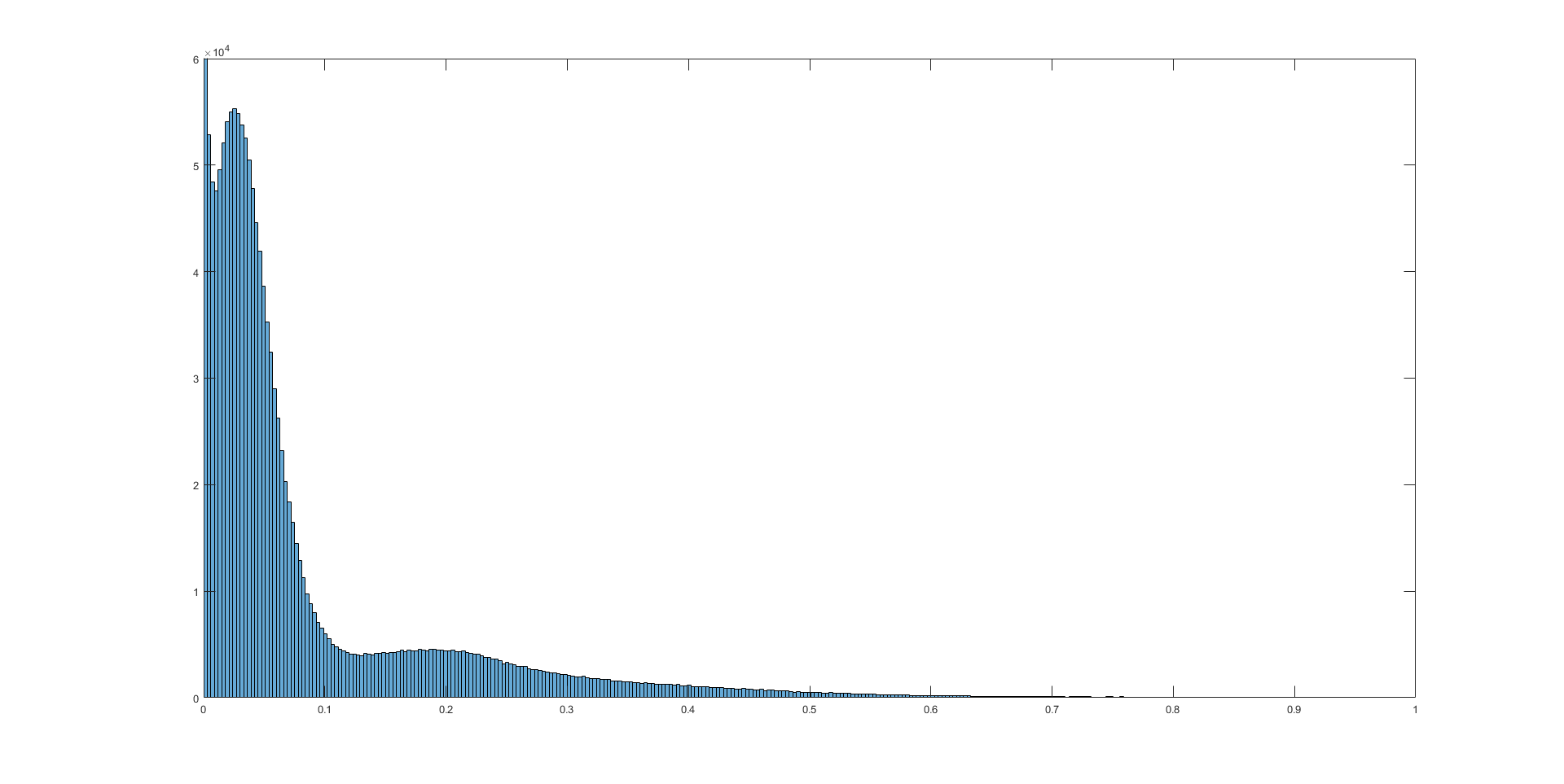}}
   \subfigure[\scriptsize GM]    {\label{fig1f} \includegraphics[width=0.18\textwidth]{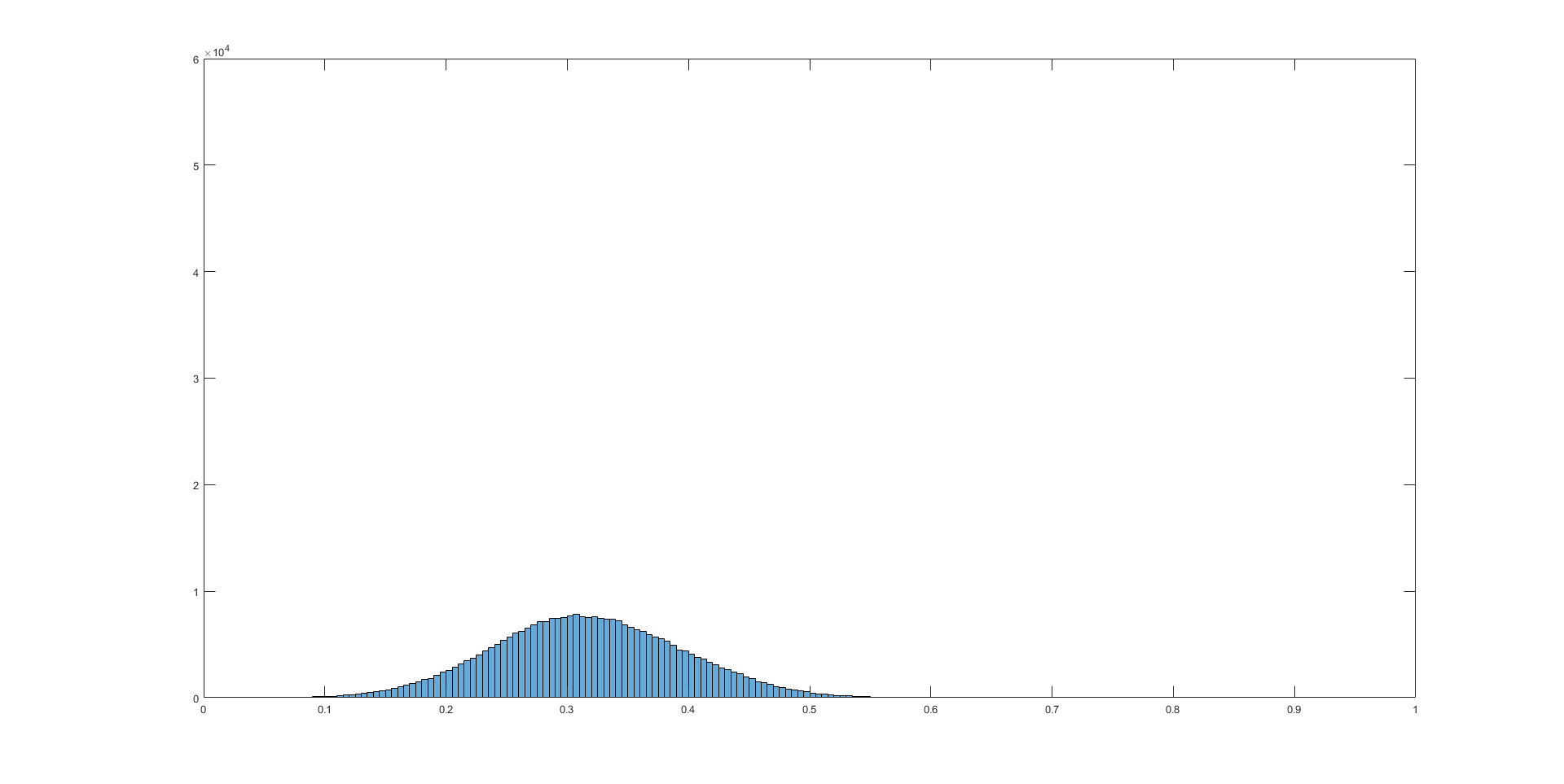}}
   \subfigure[\scriptsize WM]    {\label{fig1g} \includegraphics[width=0.18\textwidth]{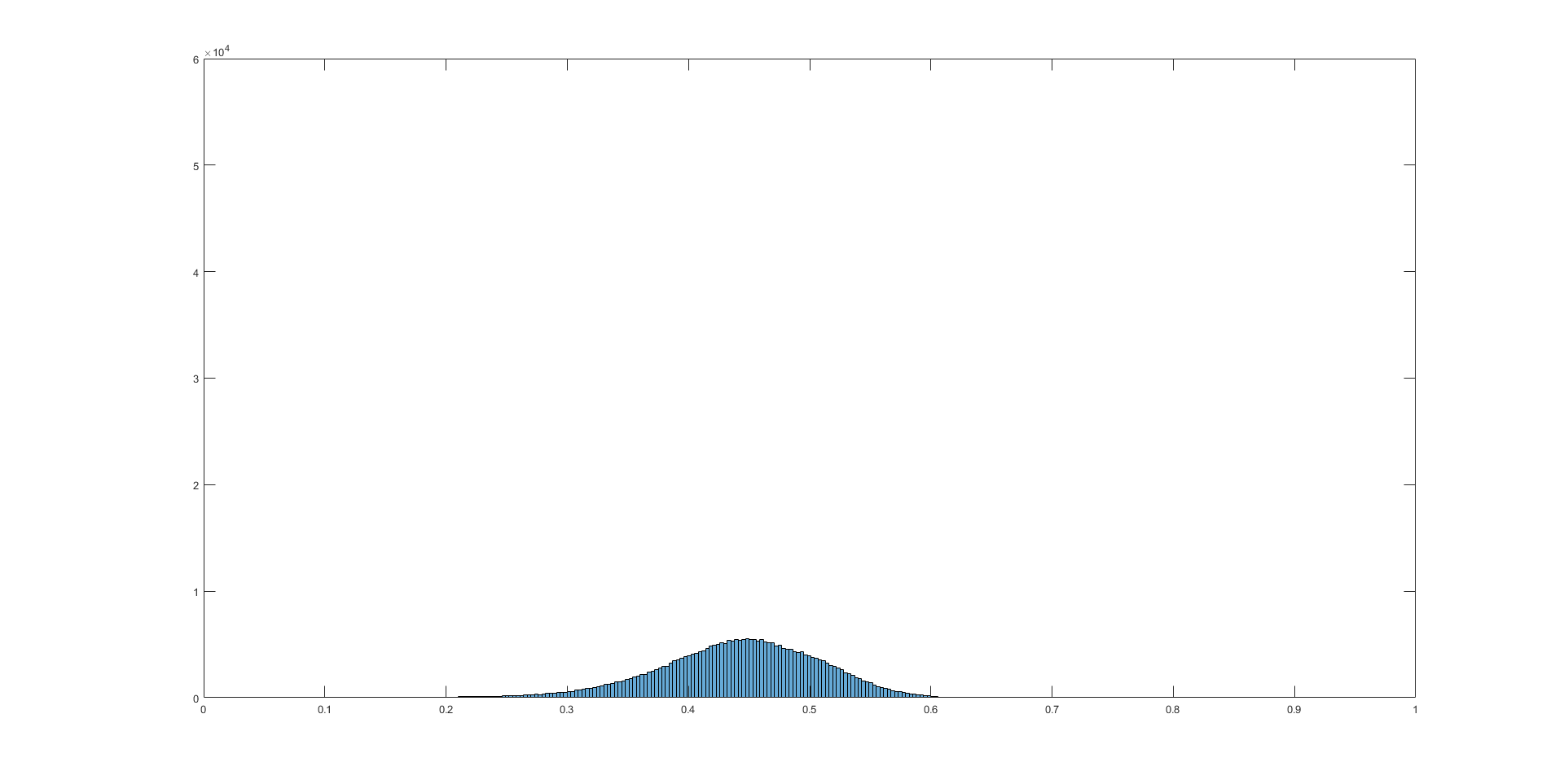}}
   \subfigure[\scriptsize CSF]   {\label{fig1h} \includegraphics[width=0.18\textwidth]{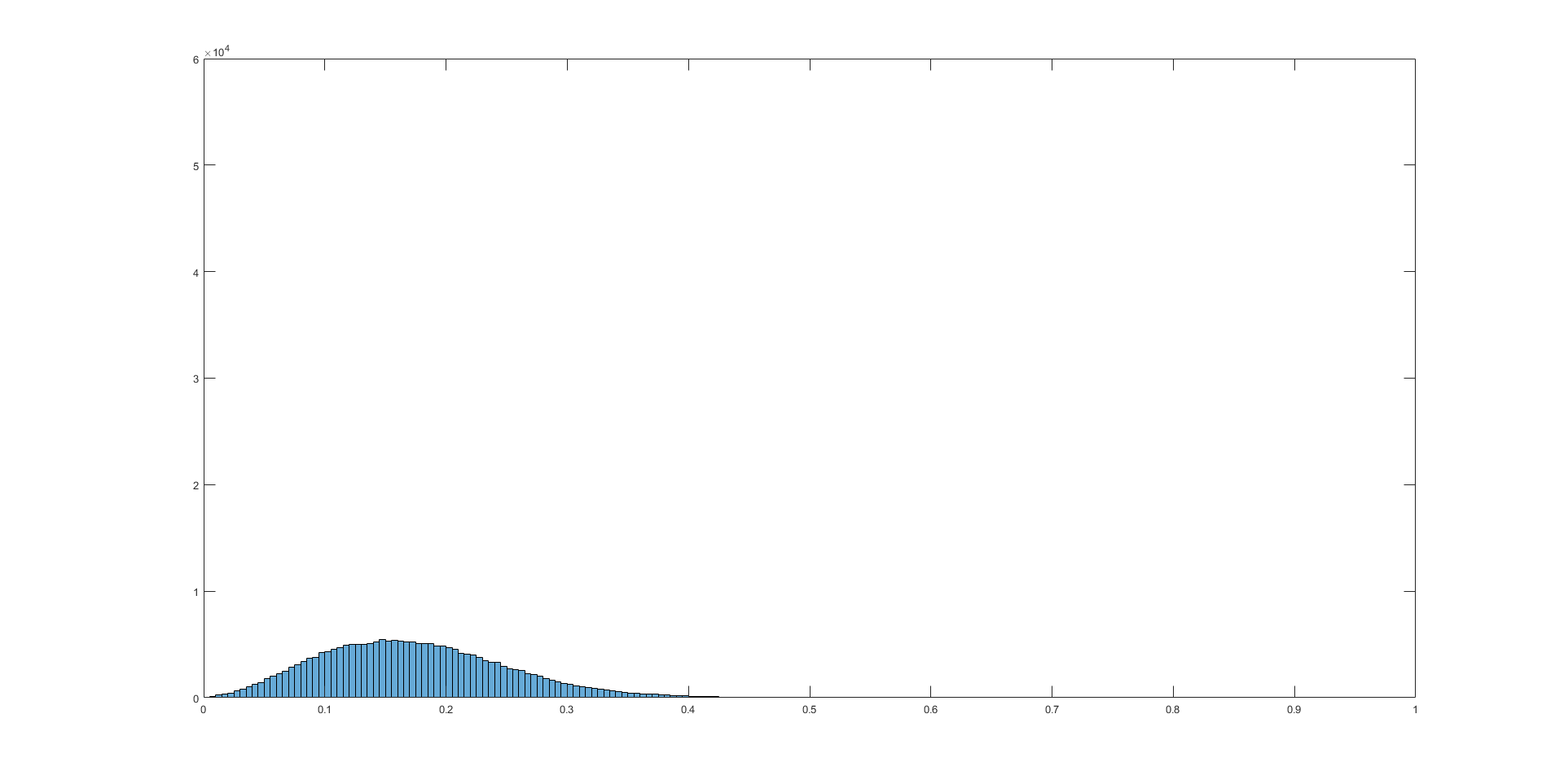}}\\
   \subfigure[\scriptsize Whole] {\label{fig1i} \includegraphics[width=0.18\textwidth]{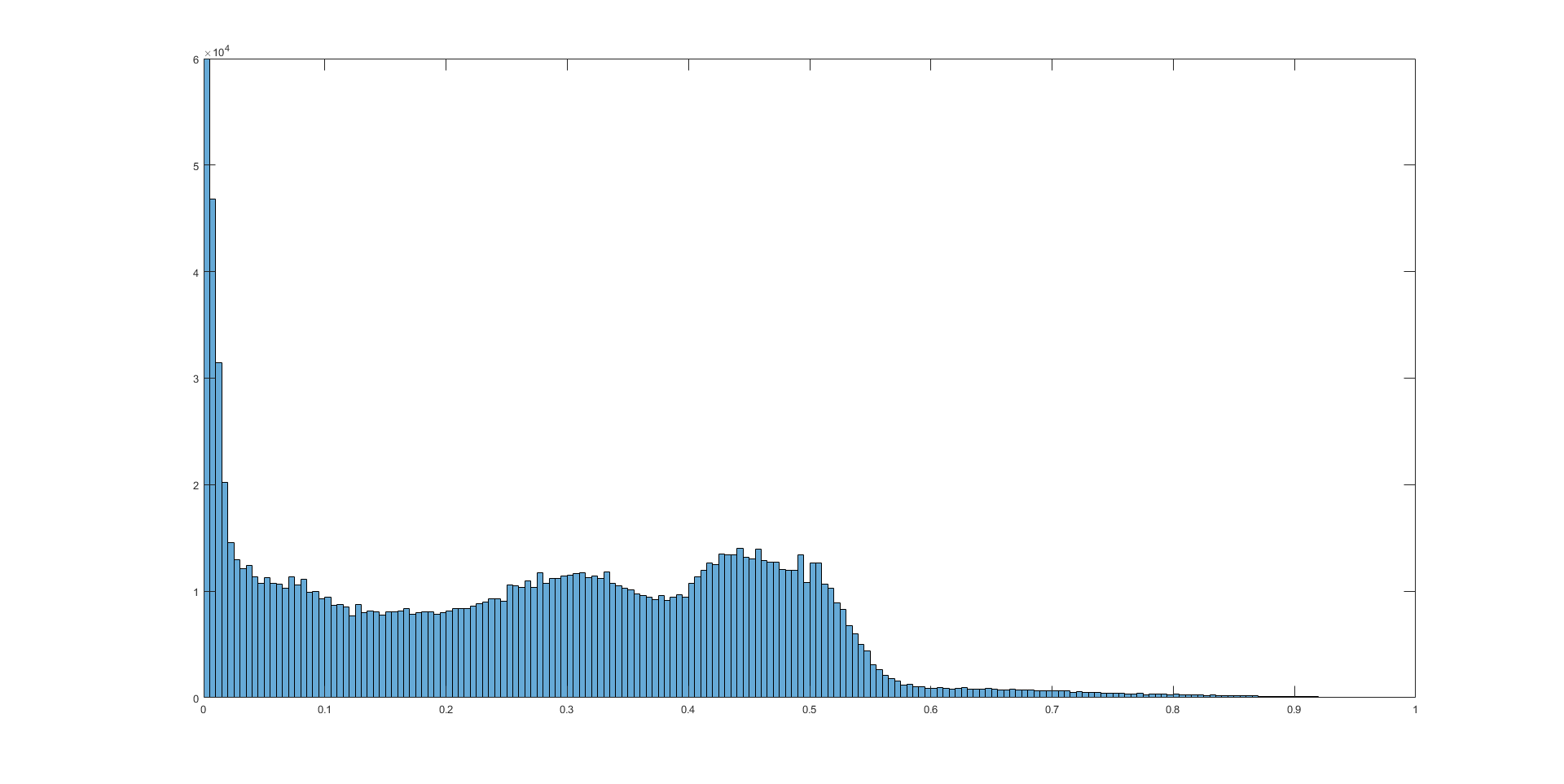}}
   \subfigure[\scriptsize BG]    {\label{fig1j} \includegraphics[width=0.18\textwidth]{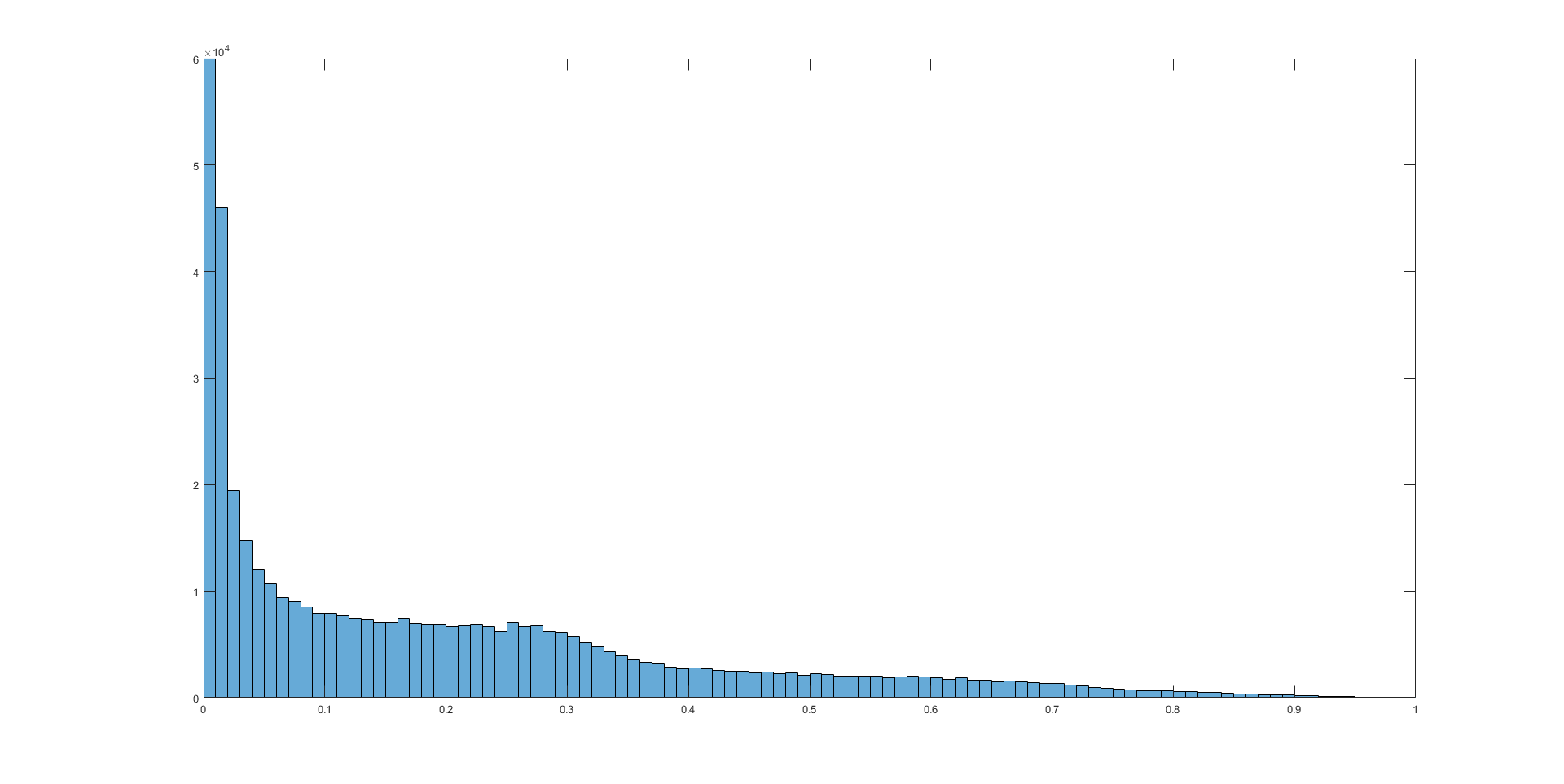}}
   \subfigure[\scriptsize GM]    {\label{fig1k} \includegraphics[width=0.18\textwidth]{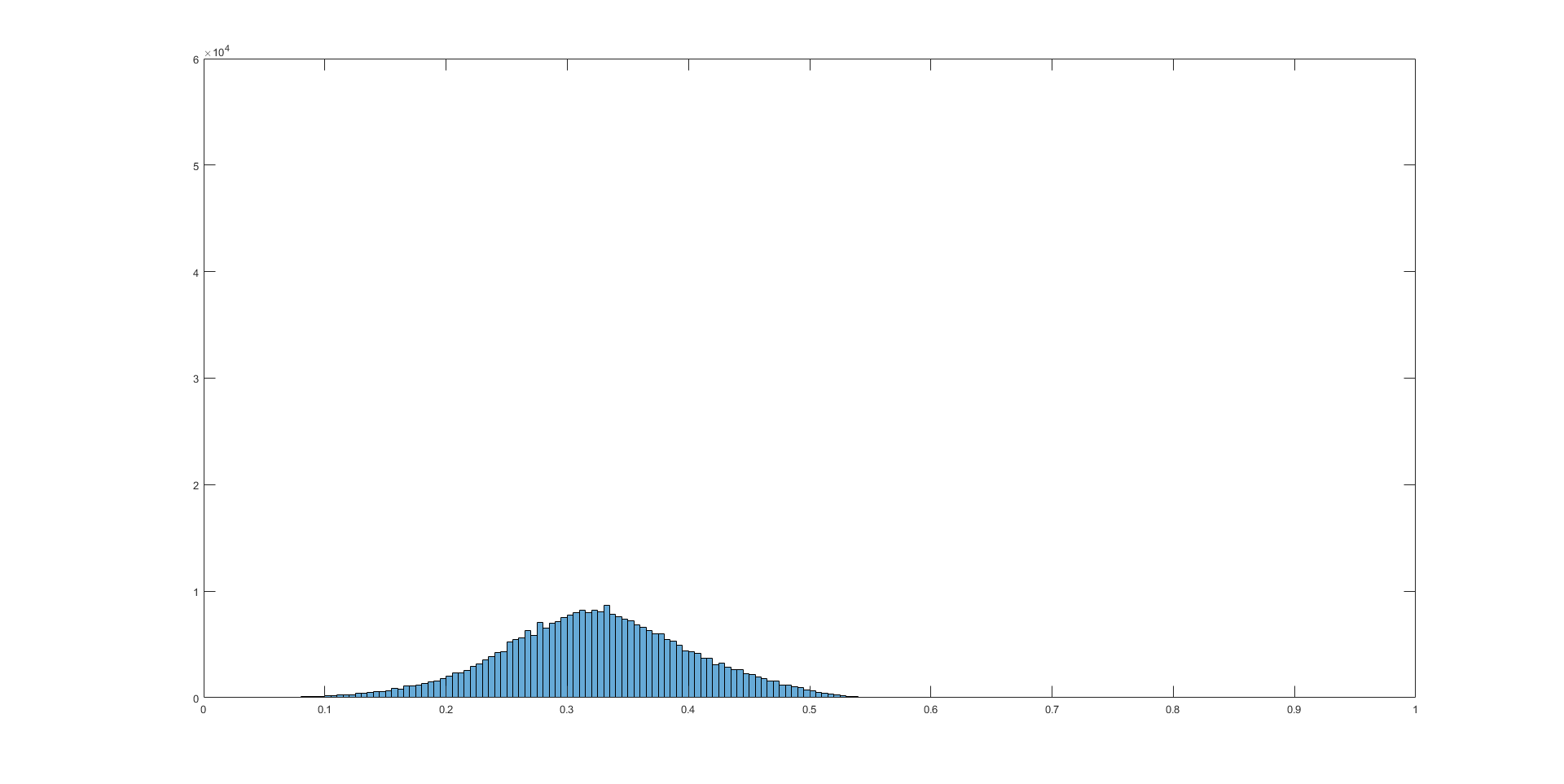}}
   \subfigure[\scriptsize WM]    {\label{fig1l} \includegraphics[width=0.18\textwidth]{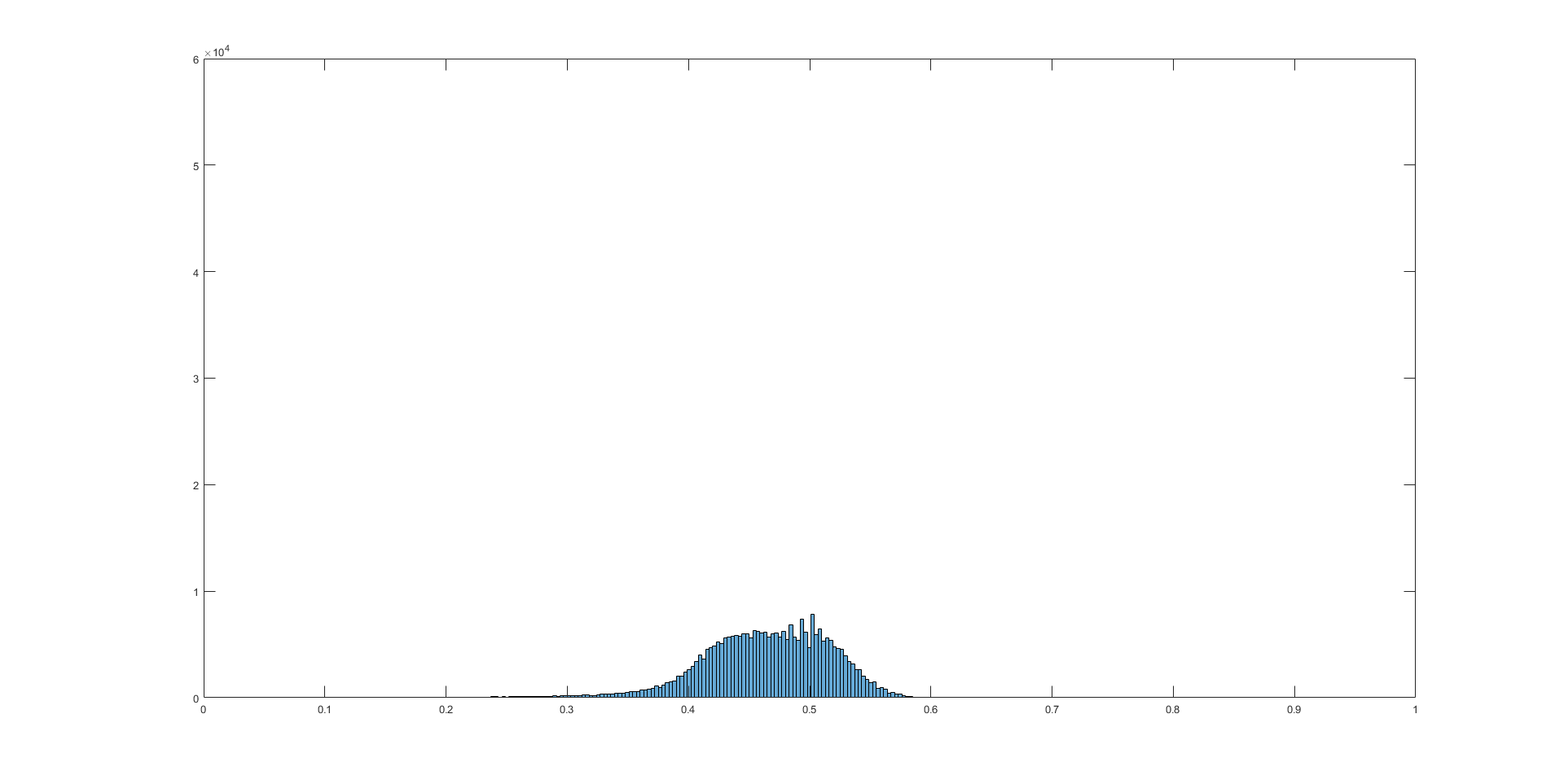}}
   \subfigure[\scriptsize CSF]   {\label{fig1m} \includegraphics[width=0.18\textwidth]{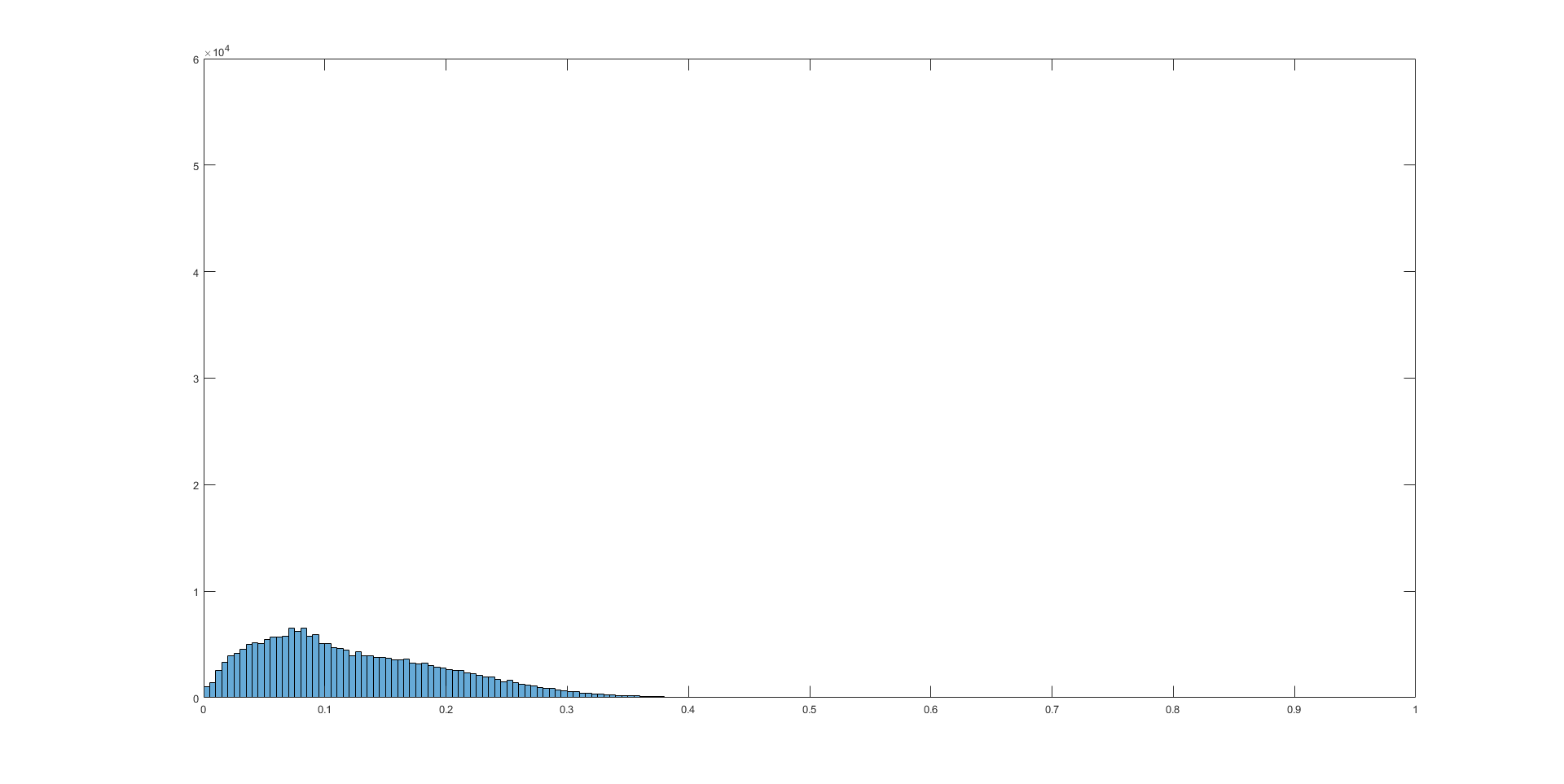}}
   \caption{A full-sampled MR image in Figure \ref{fig1a}, its under-sampled counterpart in Figure \ref{fig1b} and segmentation labels in Figure \ref{fig1c}. We plot the histograms of under-sampled MRI (second row) and full-sampled MRI (third row) on training MRI datasets.}
\label {fig1}
\end{center}
\end{figure}

Magnetic resonance imaging (MRI) is a medical imaging technique used in radiology to produce the anatomical images in human body with the advantages of low radiation, high resolution in soft tissues and multiple imaging modalities. However, the major limitation in MRI is the slow imaging speed which causes motion artifacts \cite{1} when the imaging subject moves consciously or unconsciously. The high resolution in k-t space is also difficult to be achieved in dynamic MRI because of long imaging period \cite{2}. Thus compressed sensing technique is introduced to accelerate the MRI by measuring less k-space samples called compressed sensing MRI (CS-MRI) \cite{3}. The CS-MRI is a classic inverse problem in computation imaging requiring proper regularization for accurate reconstruction.

\vspace{-10mm}

The standard CS-MRI can be formulated as
\begin{equation}\label{eq1}
\hat x = \mathop {\arg \min }\limits_x \left\| {{F_u}x - y} \right\|_2^2 + \sum\limits_i {{\alpha _i}{\Psi _i}\left( x \right)},
\end{equation}
where $x \in {C^{P \times 1}}$ is the complex-valued MR image to be reconstructed, $F_u\in {C^{M \times P}}$ is the under-sampled Fourier operator and $y\in {C^{M \times 1}}$ ($M\ll P$) are the k-space measurements by the MRI machine. The first data fidelity term ensures consistency between the Fourier coefficients of the reconstructed image and the measured k-space data, while the second prior term regularizes the reconstruction to encourage certain image properties such as sparsity in a transform domain.

In conventional CS-MRI methods, the sparse and nonlocal are common priors for the inverse recovery in situ, which brings three limitations: (1) The common complex patterns hiding massive MRI datasets are overlooked in the capacity-limited ``shallow" prior \cite{4}. (2) The sparse or nonlocal regularization lacks semantic representation ability, which is difficult to distinguish between the image structure details and structural artifacts brought by under-sampling. (3) The optimization for conventional priors requires tons of iterations to reach the convergence, which brings long reconstruction time consumption \cite{5}.

Recently, the deep neural network models are introduced in the field of CS-MRI to overcome the above limitations. where the information from massive training MRI datasets can be encoded in the network architecture in training phase with large model capacity. Once the network is well-trained, the forward reconstruction for test MRI data is much faster compared with methods based on conventional sparse priors because no iteration is required. More importantly, the deep neural network models enjoy the benefit of modeling the semantic information in the image, providing an appropriate approach to integrate information for different visual tasks, however, which is rarely considered in the existing models for inverse problem, leaving high-level supervision information poorly utilized, causing negative effect on the later automatic analysis phase.

We take segmentation information for example to prove the benefits of introducing high-level supervision information into reconstruction. Usually different tissues in the MR image not only have different diagnostic information, but also show different statistical properties. In Figure \ref{fig1a} and Figure \ref{fig1b}, we show a full-sampled and corresponding under-sampled T1-weighted brain MR image which contains three different labeled tissues: gray matter (GM), white matter (WM) and cerebrospinal fluid (CSF). The corresponding GM, WM and CSF labels are shown in green, yellow and red in the segmentation label map in Figure \ref{fig1c}. Clearly, different regions show different intensity scales. To further quantify this phenomenon, we give the statistical histograms of the three tissues, back ground (BG) and the whole images of the under-sampled/full-sampled MRI data in the second/third row in Figure \ref{fig1} on all the training MRI data. We observe each of the GM, WM and CSF tissues has simple single-mode distribution on the full-sampled and under-sampled MRI data. Since the deep neural network usually learns the function mapping from the under-sampled MR images to their full-sampled counterparts. The function mapping can be significantly simplified by learning the corresponding relations between the single-mode distributions. However, the distributions of the whole under-sampled and full-sampled MRI in Figure \ref{fig1d} and \ref{fig1i} are much more complicated, making the learning of function mapping more difficult.

In this paper, we propose a segmentation-aware deep fusion network (SADFN) architecture for compressed sensing MRI to fuse the semantic supervision information in the different depth from the segmentation label and propagate the semantic features to each layer in the reconstruction network. The main contribution can be summarized as follows:
\begin{itemize}
  \item The proposed SADFN model can effectively fuse the information from tasks and depths in different levels. Both the MRI reconstruction and segmentation accuracies are significantly improved under the proposed framework.
  \item The semantic information from the segmentation network is provided to reconstruction network using a feature fusion strategy, helping the reconstruction network be aware of
      the content it reconstructs and simplifying the function mapping.
  \item We adopt the multilayer feature aggregation to effectively collect and extract the information from different depth in the segmentation network.
\end{itemize}

\section{Related Work}
\label{ReWo}

\subsection{Compressed Sensing MRI}
\label{CSMRI}

In the study of CS-MRI, the researches focus on proposing appropriate regularization. In the pioneer work SparseMRI \cite{3}, the fixed transform operator wavelets and total variation is adopted for regularization in Equation \ref{eq1}. More methods \cite{6,7,8} are proposed to address the same objective function efficiently. The variants of wavelet are proposed to exploit the geometric information in MR images adaptively in \cite{9,10,11}. Dictionary learning techniques are also utilized in situ to model the MR images adaptively \cite{5,12,13}. Nonlocal prior also can be introduced as regularizator \cite{14} or combined with sparse prior in \cite{10}.

Recently, the deep neural network models are introduced in CS-MRI. A vanilla deep convolutional neural network (CNN) is used to learn the function mapping from the zero-filled MR images to the full-sampled MR images \cite{15}. Furthermore, a modified U-Net architecture is utilized to learn the residual mapping in \cite{17}. The above deep-based CS-MRI models overlooks the accurate information on the sampled positions in the compressive measurements. In \cite{4}, a deep cascaded CNN (DC-CNN) is proposed to cascade several basic blocks to learn the mapping with each block containing the nonlinear convolution layers and a nonadjustable data fidelity layer. In data fidelity layers, the reconstructed MR images are corrected by the accurate k-space samples. Despite the state-of-the-art reconstruction quality has been achieved using the DC-CNN model, the high-level supervision information from the manual labels in MRI datasets hasn't been taken into consideration, still leaving room for further improvement on model performance.

\subsection{MR Image Segmentation}
\label{MRIS}

With the segmentation labels in MRI datasets, different models are proposed to learn to automatically segment the MR images into different tissues from the test set. Compared with conventional segmentation methods based on manually designed features, the deep neural network models can extract image features automatically, leading to better segmentation performance. Recently, the U-shaped network called U-Net trained in end-to-end and pixel-to-pixel manner is proposed in \cite{18}, which can take the input of arbitrary size and produce the output of the same size, achieving the state-of-the-art medical image segmentation accuracy and computational efficiency. Its variant where the 2D operations are replaced with 3D ones is proposed in \cite{19} called 3D U-Net. The residual learning is also utilized in the segmentation model in \cite{20}. The recurrent neural network can efficiently model the relation among different frames in the volumetric MR data can introduced in the medical image segmentation \cite{21,22}. Throughout the paper, we use the classic 2D U-Net for single-frame MRI segmentation for the single-frame MRI reconstruction, and the proposed model can be easily extended to volumetric MRI data.

\subsection{Multilayer Feature Aggregation}
\label{MLFA}

The works \cite{23} on visualization of deep CNN has revealed the feature maps at different layers describe the image in different scales and views. In the conventional deep neural network models, the output is produced based on the deep layers or even the last layer of the model, leaving the features in lower layers containing information from different scales underemphasized. In the field of salient object detection, the multilayer feature aggregation is a popular approach to integrate information from different layers in the network \cite{24,25,26}.

\subsection{High-level Information Guidance for Low-level Tasks}
\label{CAM}

Recently, some works are devoted to combining the low-level task with tasks in higher levels. In the work of \cite{27}, a well pre-trained segmentation network is cascaded behind a denoising network, then the loss functions for both segmentation and denoising are optimized to train the denoising network without adjusting the parameters in segmentation network. With this model, the denoising network produces the denoised images with higher segmentation accuracy using automatic segmentation network at the expense of limited improvement in restoration accuracy or even degradation. In the AOD-Net \cite{28}, the well-trained dehaze model is jointly optimized with a faster R-CNN, resulting better detection and recognition results.

\section{The Proposed Architecture}
\label{TheproArchi}

\begin{figure}[htb!]
\begin{center}
   {\includegraphics[width=1\textwidth]{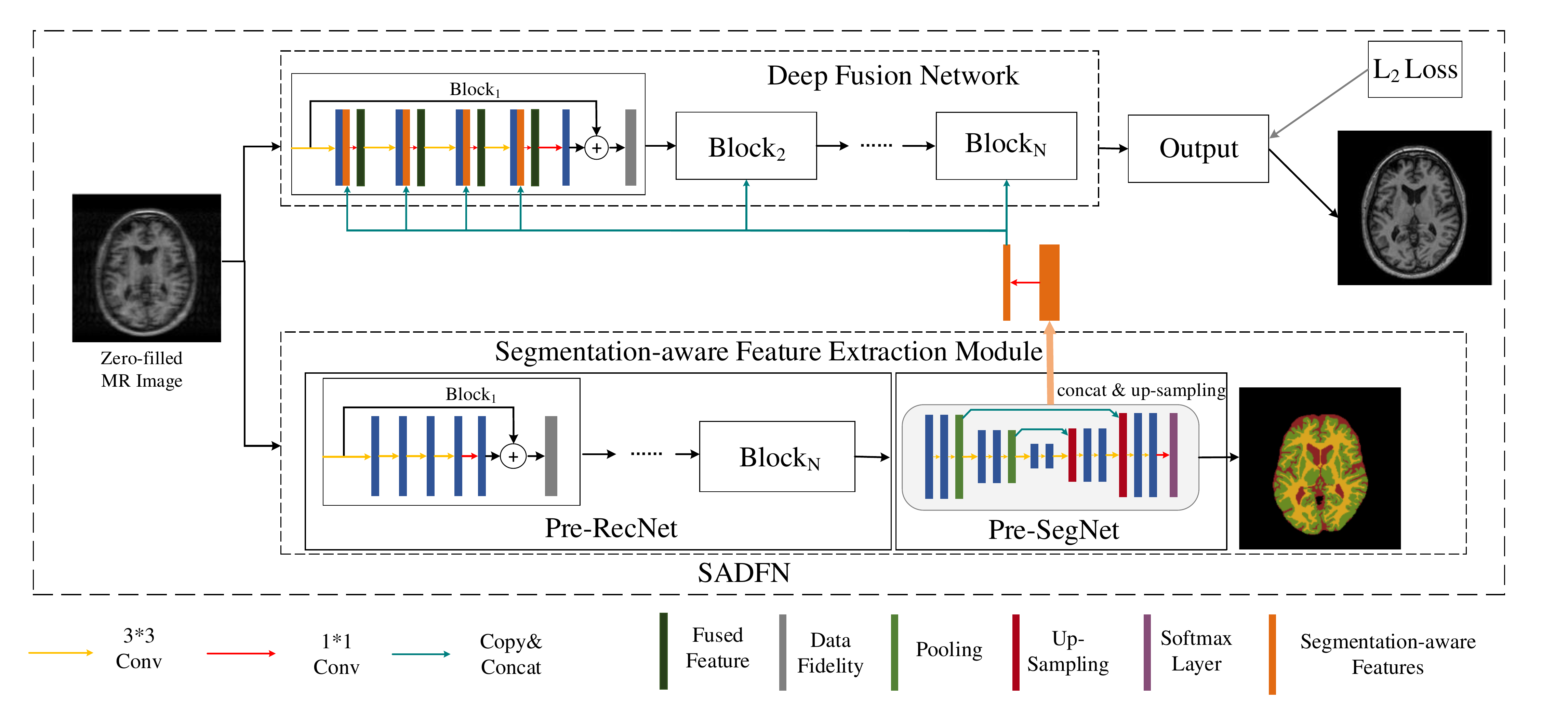}}
   \caption{The network architecture of SADFN model.}
\label {fig2}
\end{center}
\end{figure}

To incorporate the information from segmentation label into the MRI reconstruction, we proposed the segmentation-aware deep fusion network (SADFN). The network architecture is shown in Figure \ref{fig2}. The reconstruction network and segmentation network are first pre-trained. Then a segmentation-aware feature extraction module is designed to provide features with rich segmentation information to reconstruction network using a feature fusion strategy.

\subsection{The Pre-trained MRI Reconstruction Network}
\label{BaselineRec}

As we introduced above, the DC-CNN architecture achieves the state-of-the-art performance in reconstruction accuracy and computational efficiency. We train a DC-CNN network with $N$ cascaded blocks. Each block contains several nonlinear convolutional layer and a data fidelity layer. The details of each block in the DC-CNN architecture is shown in Table \ref{RecNet}. In the data fidelity layer, the input reconstructed MR image produced by the previous convolutional layer is first transformed into frequency domain. Then the accurate samples in under-sampled k-space measurements directly replace the corresponding positions in the Fourier coefficients. The corrected Fourier coefficients are transformed back to MR image as the output by the data fidelity layer. The details can also be found in \cite{4}. Note the identity function is used in last convolutional layer to admit the negative values because of the global residual learning in the blocks. We also refer to the DC-CNN architecture as Pre-RecNet for simplicity. The Pre-RecNet with $N$ blocks are called Pre-RecNet$_N$.
\begin{table}[]
\centering
\caption{The parameter setting of a block in the Pre-RecNet.}
\label{RecNet}
\begin{tabular}{|c|c|c|c|c|c|c|}
\hline
Layer         & Input      & Filter Size & Stride & Number of Filters & Activation & Output     \\ \hline
Conv$_1$         & 240*240    & 3*3         & 1      & 32                & ReLU       & 240*240*32 \\ \hline
Conv$_2$         & 240*240*32 & 3*3         & 1      & 32                & ReLU       & 240*240*32 \\ \hline
Conv$_3$         & 240*240*32 & 3*3         & 1      & 32                & ReLU       & 240*240*32 \\ \hline
Conv$_4$         & 240*240*32 & 3*3         & 1      & 32                & ReLU       & 240*240*32 \\ \hline
Conv$_5$         & 240*240*32 & 3*3         & 1      & 1                 & Linear     & 240*240    \\ \hline
Data Fidelity & 240*240    & N/A         & N/A    & N/A               & N/A        & 240*240    \\ \hline
\end{tabular}
\end{table}
We train the Pre-RecNet$_N$ using the under-sampled and full-sampled training data pairs by minimizing the following Euclidean loss function
\begin{equation}\label{eq2}
{\cal L_{{\rm{Rec}}}}\left( {{y_i},x_i^{fs};{\theta _r}} \right) = \frac{1}{{{L_r}}}\sum\limits_{i = 1}^{{L_r}} {\left\| {x_i^{fs} - {f_{{\theta _r}}}\left( {F_u^H{y_i}} \right)} \right\|_2^2}.
\end{equation}
Where the ${x_i^{fs}}$ is the full-sampled MR image, $y_i$ is the under-sampled k-space measurements in the training batch. ${\theta _r}$ denotes the network parameter and $L_r$ is the number of MRI data in the training batch.

\subsection{The MRI Segmentation Network}
\label{TheMRISeg}

To fully utilize the segmentation supervision information, we train a automatic segmentation network. We adopt the popular U-Net architecture as the segmentation model called Pre-SegNet. The parameter setting of the Pre-SegNet is shown in Table \ref{SegNet}.
\begin{table}[]
\centering
\scriptsize
\caption{The parameter setting of the Pre-SegNet.}
\label{SegNet}
\begin{tabular}{|c|c|c|c|c|c|c|}
\hline
Layer        & Input       & Filter Size & Stride & Number of Filters & Activation & Output          \\ \hline
Conv$_1$        & 240*240     & 3*3         & 1      & 32                & ReLU       & 240*240*32      \\ \hline
Conv$_2$       & 240*240*32  & 3*3         & 1      & 32                & ReLU       & 240*240*32      \\ \hline
Max Pooling$_1$ & 240*240*32  & N/A         & 2      & N/A               & N/A        & 120*120*32      \\ \hline
Conv$_3$        & 120*120*32  & 3*3         & 1      & 64                & ReLU       & 120*120*64      \\ \hline
Conv$_4$        & 120*120*64  & 3*3         & 1      & 64                 & ReLU       & 120*120*64      \\ \hline
Max Pooling$_2$ & 120*120*64  & N/A         & 2      & N/A               & N/A        & 60*60*64        \\ \hline
Conv$_5$        & 60*60*64    & 3*3         & 1      & 128               & ReLU       & 60*60*128       \\ \hline
Conv$_6$        & 60*60*128   & 3*3         & 1      & 128                & ReLU       & 60*60*128        \\ \hline
Deconv$_1$      & 60*60*128    & 3*3         & 1      & 64                & ReLU       & 120*120*64 \\ \hline
Conv$_7$        & 120*120*(64+64) & 3*3         & 1      & 64                & ReLU       & 120*120*64      \\ \hline
Conv$_8$        & 120*120*64  & 3*3         & 1      & 64                & ReLU       & 120*120*64      \\ \hline
Deconv$_2$      & 120*120*64  & 3*3         & 1      & 32                & ReLU       & 240*240*32 \\ \hline
Conv$_9$       & 240*240*(32+32)  & 3*3         & 1      & 32                & ReLU       & 240*240*32      \\ \hline
Conv$_{10}$       & 240*240*32  & 3*3         & 1      & 32                & ReLU       & 240*240*32      \\ \hline
Conv$_{11}$       & 240*240*32  & 3*3         & 1      & 4                 & Linear       & 240*240*4         \\ \hline
Softmax      & 240*240*4     & N/A         & N/A    & N/A               & N/A        & 240*240         \\ \hline
\end{tabular}
\end{table}
The pooling operation can help the network extract the image features in different scales and the symmetric concatenation is utilized to propagate the low-layer features to high layers directly, providing accurate localization. We train the Pre-SegNet using the full-sampled MR images and their corresponding segmentation labels as training data pairs by minimizing the following pixel-wise cross-entropy loss function
\begin{equation}\label{eq3}
{{\cal L}_{{\rm{Seg}}}}\left( {x_i^{fs},t_i^{gt};{\theta _s}} \right) =  - \sum\limits_{i = 1}^{{L_s}} {\sum\limits_{j = 1}^R {\sum\limits_{c = 1}^C {t_{ijc}^{gt}} } } \ln {t_{ijc}}.
\end{equation}
Where the ${t_i^{gt}}$ is the segmentation label in the training batch and $t_i$ is the corresponding segmentation result produced by Pre-SegNet. ${\theta _s}$ denotes the network parameter and $L_s$ is the number of MRI data in the training batch. $C$ denotes the the number of classes of the label. Taking the brain segmentation for example \cite{29}, the brain tissues can be classified into white matter, gray matter, cerebrospinal fluid and background. Thus $C$ is $4$ for segmentation.

\subsection{Deep Fusion Network}
\label{DFN}

With the well-trained Pre-RecNet and Pre-SegNet, we can construct the segmentation-aware deep fusion network with $N$ blocks (SADFN$_N$) by integrating the features from the Pre-RecNet and Pre-SegNet, which involving a cross-layer multilayer feature aggregation strategy and a cross-task feature fusion strategy.

\subsubsection{Segmentation-aware Feature Extraction Module}
\label{MLFAinSeg}

As we discussed in the related work section, the multilayer feature aggregation can be used to fuse the information from layers in different depth. Here we extract the feature maps from the output of the Conv$_1$, Conv$_2$, Conv$_3$, Conv$_4$, Conv$_5$, Conv$_6$, Conv$_7$, Conv$_8$, Conv$_9$, Conv$_{10}$ and concatenate them into a single ``thick" feature map tensor. Note the smaller size feature maps are up-sampled using bilinear interpolation to the same size of features from the Pre-RecNet$_N$. Then the ``thick" feature maps of the size $240*240*640$ (32+32+64+64+128+128+64+64+32+32) are further compressed into a ``thin" feature tensor of the size $240*240*32$ via the $1\times1$ convolution with ReLU as activation function.

\subsubsection{The Feature Fusion cross Tasks}
\label{TFFinRec}

The compressed feature tensor obtained by the multilayer feature aggregation strategy contains the supervision information from the Pre-SegNet. We concatenate the feature tensor of the size $240*240*32$ with the feature maps of the size $240*240*32$ output by convolutional layers in the Pre-RecNet as shown in Figure \ref{fig2}. Then the concatenated features of the size $240*240*64$ are further compressed into a feature tensor of the size $240*240*32$ via $1\times1$ convolution with ReLU activation function. The compressed feature tensor from multilayer feature aggregation is concatenated to the first four convolutional layers in each Pre-RecNet block, the supervision information from segmentation can guide the reconstruction in different depth. Also, in the Figure \ref{fig2}, the feature fusion strategy is also utilized in each block of the Pre-RecNet.

\begin{figure}[htb!]
\begin{center}
   \subfigure[\scriptsize Seg Labels] {\label{fig3a} \includegraphics[width=0.235\textwidth]{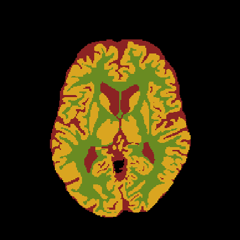}}
   \subfigure[\scriptsize Feature ($2^{nd}$ Layer)]  {\label{fig3b} \includegraphics[width=0.235\textwidth]{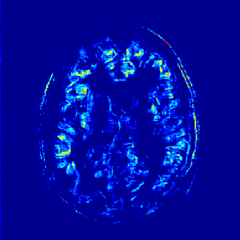}}
   \subfigure[\scriptsize Feature ($4^{th}$ Layer)] {\label{fig3c} \includegraphics[width=0.235\textwidth]{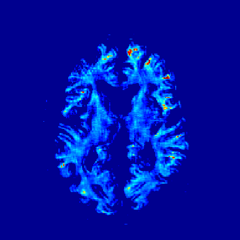}}
   \subfigure[\scriptsize Feature ($4^{th}$ Layer)] {\label{fig3d} \includegraphics[width=0.235\textwidth]{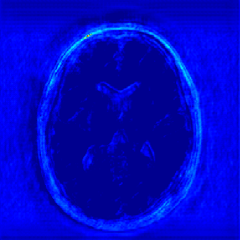}}
   \caption{The selected feature maps from the feature tensors produced by the feature fusion in the deep fusion network.}
\label {fig3}
\end{center}
\end{figure}

To prove the supervision information is effectively fused into the reconstruction, we give some feature maps in the fused feature tensor yielded by the $1\times1$ convolution in Figure \ref{fig3}. In Figure \ref{fig3a} we show the segmentation label of a certain MRI data. In Figure \ref{fig3b}, Figure \ref{fig3c} and Figure \ref{fig3d}, we visualize the feature maps selected from the fused feature tensors in the second layer and fourth layer. We observe the feature maps show clear segmentation information, while no such feature maps are observed in the Pre-RecNet$_N$ model.

\vspace{-10mm}

\subsubsection{The Fine-tuning Strategy}
\label{TheFineTune}

We further fine-tune the resulting architecture. Given a zero-filled MR image in the training dataset, a corresponding high-quality MR image can be yielded by the Pre-RecNet$_N$ in Section \ref{BaselineRec}. Then the MR image is sent to the Pre-SegNet to extract the segmentation features, which are then utilized for the multilayer feature aggregation in Pre-SegNet and feature fusion. Meanwhile, the zero-filled MR image is also input to the deep fusion network. The $\ell_2$ Euclidean distance between the output reconstructed MR image and the corresponding full-sampled MR image in the training dataset is minimized. During the optimization, the parameters in the Pre-RecNet$_N$ and Pre-SegNet are kept fixed, while we only adjust the parameters in the deep fusion network and the $1\times1$ convolutions in multilayer feature aggregation.

\vspace{-2mm}

\section{Experiments}

\subsection{Datasets}

We train and test our SADFN model on MRBrainS datasets from Grand Challenge on MR Brain Image Segmentation(MRBrainS) Benchmark \cite{29}. The datasets provides well-aligned multiple modalities MRI including T1, T1-IR and T2-FLAIR with segmentation labels by human experts. For simplicity, we only use the T1 weighted MRI data. In the future work, we plan to extend the model on multi-modalites MRI imaging. Total 5 scans are provided public segmentation labels. We randomly choose four scans for training containing total 192 slices. The training MR images are of size 240$\times$240. We use the left MRI scan for testing the model performance containing total 48 slices.

\subsection{Implementation Details}

We train and test the algorithm on Tensorflow for the Python environment on a NVIDIA Geforce GTX 1080Ti with 11GB GPU memory and Intel Xeon CPU E5-2683 at 2.00GHz. The detailed network architectures for Pre-RecNet, Pre-SegNet and SADFN have been introduced in previous section.

The ADAM is used as the optimizer. We train the Pre-RecNet for 32000 iterations using a batch containing four under-sampled and their corresponding full-sampled MR images as training pairs in Equation \ref{eq2}. The Pre-SegNet is also pre-trained for 32000 iterations using a batch containing 16 randomly cropped fully-sampled 128${\times}$128 patches and their segmentation labels. Again, we note that during the fine-tuning of the SADFN model, compressed feature tensor is yielded by multilayer features aggregation (MLFA) and the feature tensor is propagated to the Pre-RecNet before the feature fusion in each block. The SADFN is fine-tuned 12000 iterations using the same training batchsize as the pre-training of Pre-RecNet. We select the initial learning rate to be 0.001 for pre-trained stage and 0.0001 for fine-tune stage, the first-order momentum to be 0.9 and the second momentum to be 0.999 for both stages. We adopt batch normalization (BN) in Pre-SegNet. We also adopt data augmentation for training as implemented in \cite{30}.

\subsection{Quantitative Evaluation}

We use peak signal-to-noise ratio (PSNR) and structural similarity index (SSIM) \cite{31} for the reconstruction quantitative evaluation. We adopt a 30$\%$ 1D Cartesian pattern for under-sampling. We compare the proposed SADFN$_5$ with other state-of-the-art CS-MRI models including transform learning MRI (TLMRI) \cite{12}, patch-based nonlocal operator (PANO) \cite{10}, graph-based redundant wavelet tranform (GBRWT) \cite{11}, the Pre-RecNet$_5$ (which is also the state-ot-the-art DC-CNN with 5 blocks \cite{4}). For the non-deep CS-MRI methods, we adjust the parameters to their best performance. We also compare the proposed SADFN$_5$ with the the model proposed in \cite{27}, where the pre-trained Pre-RecNet$_5$ and Pre-SegNet are cascaded during fine-tuning and only the parameters in Pre-RecNet$_5$ are adjusted for optimization. Since no name for the model is provided in the original work, we refer the model as Liu \cite{27}. Besides, we compare the proposed SADFN model with the model without the guidance of segmentation information (SADFN-WOS). For fair comparison, we design the building block of the SADFN-WOS network architecture in Table \ref{SADFN-WOS}. Note the network architecture is kept unchanged with the only difference is some feature maps in SADFN come from Pre-SegNet while all the features come from the reconstruction network in SADFN$_5$-WOS. In the model Pre-RecNet$_5$ and SADFN$_5$-WOS, no segmentation label is utilize for training, meaning the corresponding supervision information is overlooked.

\begin{table}[]
\centering
\caption{The parameter setting of a block in the SADFN-WOS model}
\label{SADFN-WOS}
\begin{tabular}{|c|c|c|c|c|c|c|}
\hline
Layer         & Input      & Filter Size & Stride & Number of Filters & Activation & Output     \\ \hline
Conv$_1$         & 240*240    & 3*3         & 1      & 64                & ReLU       & 240*240*64 \\ \hline
Conv$_2$         & 240*240*64 & 1*1         & 1      & 32                & ReLU       & 240*240*32 \\ \hline
Conv$_3$         & 240*240*32 & 3*3         & 1      & 64                & ReLU       & 240*240*64 \\ \hline
Conv$_4$         & 240*240*64 & 1*1         & 1      & 32                & ReLU       & 240*240*32 \\ \hline
Conv$_5$         & 240*240*32 & 3*3         & 1      & 64                & ReLU       & 240*240*64 \\ \hline
Conv$_6$         & 240*240*64 & 1*1         & 1      & 32                & ReLU       & 240*240*32 \\ \hline
Conv$_7$         & 240*240*32 & 3*3         & 1      & 64                & ReLU       & 240*240*64 \\ \hline
Conv$_8$         & 240*240*64 & 1*1         & 1      & 32                & ReLU       & 240*240*32 \\ \hline
Conv$_{9}$        & 240*240*32 & 3*3         & 1      & 1                 & Linear     & 240*240    \\ \hline
Data Fidelity & 240*240    & N/A         & N/A    & N/A               & N/A        & 240*240    \\ \hline
\end{tabular}
\end{table}

\begin{figure}
\begin{center}
   \subfigure[PSNR] {\label{fig4a} \includegraphics[width=0.487\textwidth]{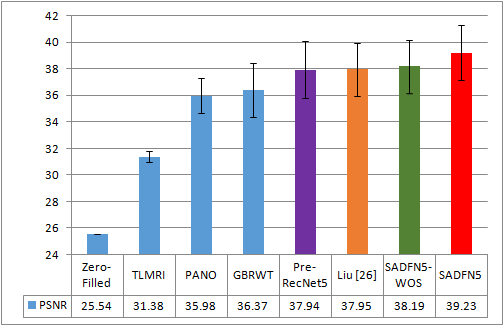}}
   \subfigure[SSIM]  {\label{fig4b} \includegraphics[width=0.487\textwidth]{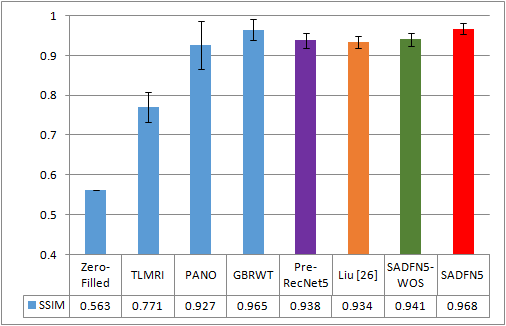}}
   \caption{The comparison in averaged PSNR and SSIM index on the test MRI data.}
\label {fig4}
\end{center}
\end{figure}

We show the objective evaluation indexes in Figure \ref{fig4}. Note the deep-based models outperform most non-deep CS-MRI models in reconstruction. We observe the proposed SADFN$_5$ model achieves the optimal performance in PSNR and SSIM indexes among the compared methods. From the standard deviation of the indexes. we note the improvement of the SADFN$_5$ is quite steady for different MRI test data. We observe the model Liu \cite{27} brings little improvement in objective evaluation indexes compared with the Pre-RecNet$_5$. We also observe the SADFN$_5$ model outperforms the comparative SADFN$_5$-WOS around 1dB in PSNR and 0.03 in SSIM in average, which proves the benefits are brought by introducing the supervision information from the segmentation labels instead of merely increasing the network size.

\subsection{Qualitative Evaluation}

\begin{figure}[htb!]
\begin{center}
   \subfigure[\scriptsize Full-sampled]                      {\label{fig5a} \includegraphics[width=0.18\textwidth]{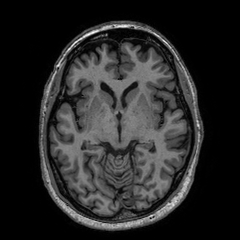}}
   \subfigure[\scriptsize Mask]                    {\label{fig5b} \includegraphics[width=0.18\textwidth]{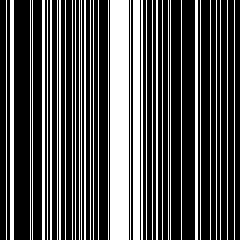}}
   \subfigure[\scriptsize Zero-filled]                      {\label{fig5c} \includegraphics[width=0.18\textwidth]{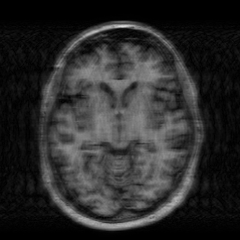}}
   \subfigure[\scriptsize TLMRI]                   {\label{fig5d} \includegraphics[width=0.18\textwidth]{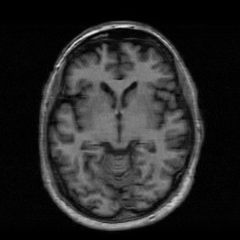}}
   \subfigure[\scriptsize PANO]                    {\label{fig5e} \includegraphics[width=0.18\textwidth]{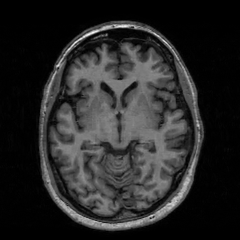}}\\
   \subfigure[\scriptsize GBRWT]                   {\label{fig5f} \includegraphics[width=0.18\textwidth]{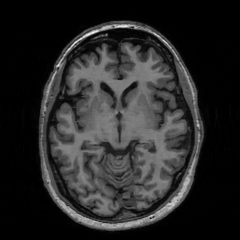}}
   \subfigure[\scriptsize Pre-RecNet$_5$]              {\label{fig5g} \includegraphics[width=0.18\textwidth]{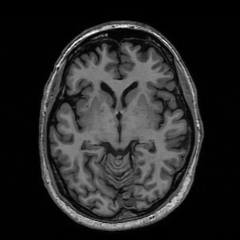}}
   \subfigure[\scriptsize Liu \cite{27}]           {\label{fig5h} \includegraphics[width=0.18\textwidth]{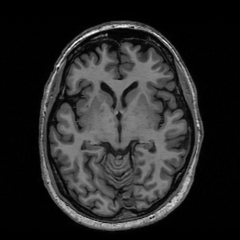}}
   \subfigure[\scriptsize SADFN$_5$-WOS]             {\label{fig5i} \includegraphics[width=0.18\textwidth]{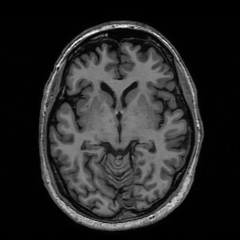}}
   \subfigure[\scriptsize SADFN$_5$]               {\label{fig5j} \includegraphics[width=0.18\textwidth]{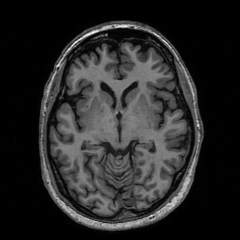}}\\
   \subfigure[\scriptsize $\Delta$ ZF]             {\label{fig5h} \includegraphics[width=0.23\textwidth]{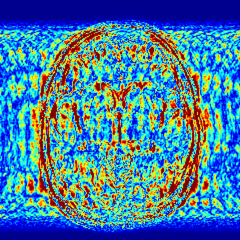}}
   \subfigure[\scriptsize $\Delta$ TLMRI]          {\label{fig5i} \includegraphics[width=0.23\textwidth]{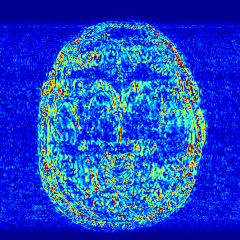}}
   \subfigure[\scriptsize $\Delta$ PANO]           {\label{fig5j} \includegraphics[width=0.23\textwidth]{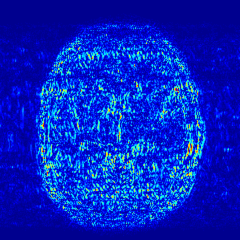}}
   \subfigure[\scriptsize $\Delta$ GBRWT]          {\label{fig5k} \includegraphics[width=0.23\textwidth]{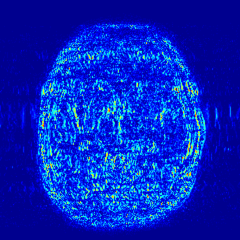}}\\
   \subfigure[\scriptsize $\Delta$ Pre-RecNet$_5$]     {\label{fig5l} \includegraphics[width=0.23\textwidth]{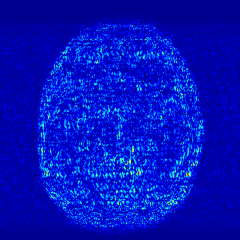}}
   \subfigure[\scriptsize $\Delta$ Liu \cite{27}]  {\label{fig5m} \includegraphics[width=0.23\textwidth]{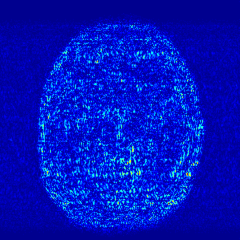}}
   \subfigure[\scriptsize $\Delta$ SADFN$_5$-WOS]    {\label{fig5n} \includegraphics[width=0.23\textwidth]{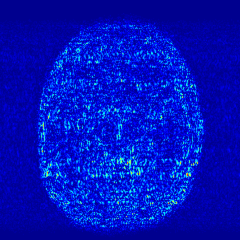}}
   \subfigure[\scriptsize $\Delta$ SADFN$_5$]      {\label{fig5o} \includegraphics[width=0.23\textwidth]{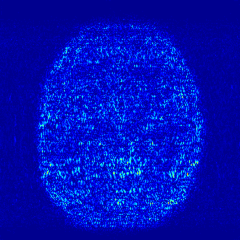}}
   \caption{The reconstruction results of zero-filled (ZF), TLMRI, PANO, GBRWT, Pre-RecNet$_5$, Liu \cite{27}, SADFN$_5$-WOS and SADFN$_5$. We also give the corresponding reconstruction error maps $\Delta$ with display ranges [0 0.12].}
\label {fig5}
\end{center}
\end{figure}

We give the qualitative reconstruction results produced by compared CS-MRI methods in Figure \ref{fig5}. We also plot the reconstruction error maps to better observe their differences. The display range for the error maps is [0 0.12]. We observe the Pre-RecNet$_5$ (DC-CNN \cite{4}) architecture, produce better reconstruction than the conventional sparse- and nonlocal- regularized CS-MRI models. The model in \cite{27} didn't brought significant improvement in reconstruction. The SADFN$_5$-WOS with larger network size also brought limited improvement. We observe the proposed SADFN$_5$ achieves much smaller reconstruction errors compared with other models, which is consistent with our observations in objective index evaluations.

\subsection{Running Time}

We compare the running time of the compared models in Table \ref{TimeCon}. As we mentioned in the Section \ref{Intro}, the CS-MRI models based on sparse or non-local regularization requires a large number of iterations, resulting slow reconstruction speed. Although the running time of the proposed SADFN model is slower than the other deep-based CS-MRI models, it achieves the state-of-the-art reconstruction accuracy, providing the best balance between running time and reconstruction quality.

\begin{table}[]
\centering
\caption{The comparison in runtime (seconds) between the compared models.}
\label{TimeCon}
\begin{tabular}{|c|c|c|c|c|c|c|c|}
\hline
                  & TLMRI  & GBRWT  & PANO  & Pre-RecNet$_5$ & Liu{[{26}]} & SADFN$_5$-WOS & SADFN$_5$ \\ \hline
Runtime & 127.67 & 100.60 & 11.37 & 0.03    & 0.03        & 0.07       & 0.06   \\ \hline
\end{tabular}
\end{table}

\section{Discussions}

\subsection{The Number of Blocks}

\begin{figure}[htb!]
\begin{center}
   \subfigure[PSNR] {\label{fig6a} \includegraphics[width=0.487\textwidth]{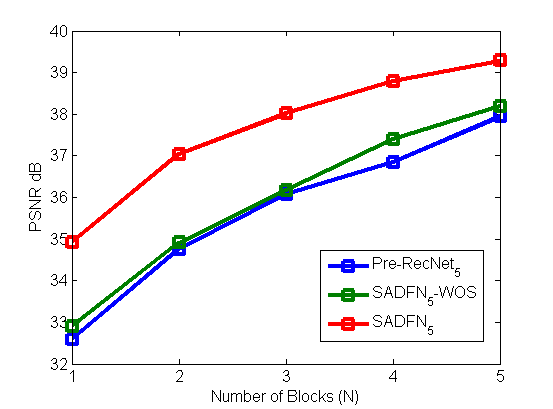}}
   \subfigure[SSIM]  {\label{fig6b} \includegraphics[width=0.487\textwidth]{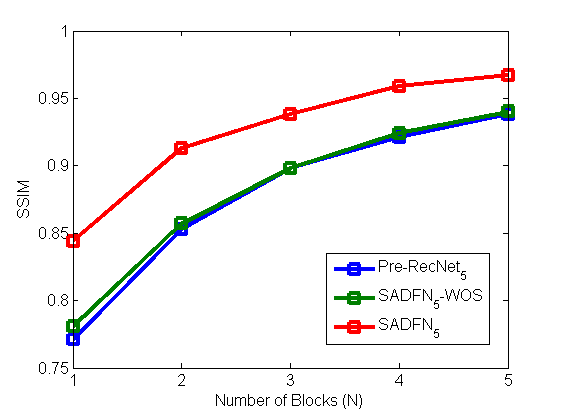}}
   \caption{The comparison in averaged PSNR and SSIM index on the test MRI data.}
\label {fig6}
\end{center}
\end{figure}

In Figure \ref{fig6}, we discuss how the model performance varies with the different number of blocks from 1 to 5 in the Pre-RecNet$_5$, SADFN$_5$-WOS and SADFN$_5$ models. As expected, the SADFN$_5$ model achieves steady improvement to large margins with different model capacity, meaning the supervision information can robustly improve the reconstruction accuracy.

\subsection{Different Under-sampling Patterns}

We also test the proposed SADFN model on the $20\%$ Random under-sampling mask shown in Figure \ref{fig5}. The SADFN$_5$ achieves the optimal performance, proving it can be well generalized on various kind of under-sampling patterns.

\begin{figure}[htb!]
\begin{center}
   \subfigure[\scriptsize Full-sampled]  {\label{fig7a} \includegraphics[width=0.18\textwidth]{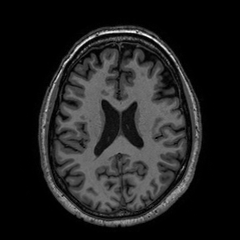}}
   \subfigure[\scriptsize Mask]  {\label{fig7b} \includegraphics[width=0.18\textwidth]{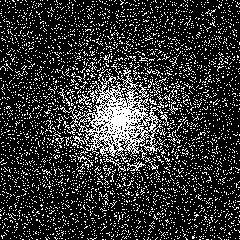}}
   \subfigure[\scriptsize GBRWT]  {\label{fig7c} \includegraphics[width=0.18\textwidth]{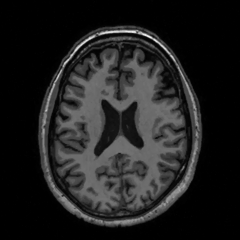}}
   \subfigure[\scriptsize Pre-RecNet$_5$]  {\label{fig7d} \includegraphics[width=0.18\textwidth]{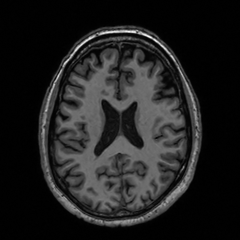}}
   \subfigure[\scriptsize SADFN$_5$]  {\label{fig7e} \includegraphics[width=0.18\textwidth]{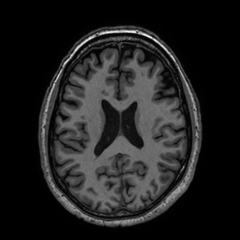}}\\
   \subfigure[\scriptsize $\Delta$ GBRWT]  {\label{fig7f} \includegraphics[width=0.3\textwidth]{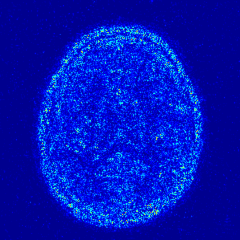}}
   \subfigure[\scriptsize $\Delta$ Pre-RecNet$_5$]  {\label{fig7g} \includegraphics[width=0.3\textwidth]{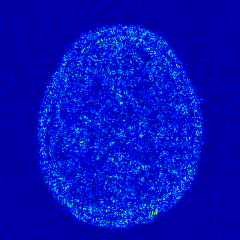}}
   \subfigure[\scriptsize $\Delta$ SADFN$_5$]  {\label{fig7h} \includegraphics[width=0.3\textwidth]{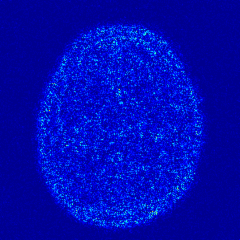}}
   \caption{The reconstruction results of zero-filled (ZF), TLMRI, PANO, GBRWT, Pre-RecNet$_5$, Liu \cite{27}, SADFN$_5$-WOS and SADFN$_5$ on the $20\%$ Random mask. We also give the corresponding reconstruction error maps $\Delta$ with display ranges [0 0.1].}
\label {fig7}
\end{center}
\end{figure}

\subsection{The Evaluation on the Segmentation Performance}

With the reconstructed MR images produced by different CS-MRI models, we input them into the pre-trained automatic segmentation models in Section \ref{TheMRISeg} to evaluate the effect of different reconstruction models on the segmentation task. We adopt the Dice Coefficient (DC), the 95th-percentile of the Hausdoff distance (HD) and the absolute volume difference (AVD) as objective evaluation indexes for segmentation as recommended in \cite{29}. The higher DC, lower HD and lower AVD values indicate better segmentation accuracy. Details on evaluation of segmentation performance can be referred to \cite{29}. The segmentation results with full-sampled MR image inputs are the performance upper bounds. We show the averaged segmentation results with compared models on the test MRI data set in Table \ref{SegPerf}. We observe the proposed SADFN$_5$ achieves the best accuracy on the segmentation task of the compared models.

\begin{table}[]
\centering
\scriptsize
\caption{The averaged DC, HD and AVD values on the test MRI data.}
\label{SegPerf}
\begin{tabular}{|c|c|c|c|c|c|c|c|c|c|}
\hline
\multirow{2}{*}{Methods} & \multicolumn{3}{c|}{GM}  & \multicolumn{3}{c|}{WM}  & \multicolumn{3}{c|}{CSF} \\ \cline{2-10}
                         & DC \%   & HD     & AVD    & DC \%  & HD     & AVD    & DC \%  & HD     & AVD    \\ \hline
ZF+Pre-SegNet                & 64.78 & 2.587 & 6.202 & 54.07 & 2.085 & 4.294 & 57.37 & 2.221 & 4.689 \\ \hline
TLMRI+Pre-SegNet             & 76.28 & 2.093 & 3.985 & 63.77 & 1.870 & 3.185 & 68.17 & 2.072 & 3.796 \\ \hline
PANO+Pre-SegNet              & 83.73 & 1.819 & 2.958 & 75.72 & 1.348 & 1.815 & 78.93 & 1.653 & 2.361 \\ \hline
GBRWT+Pre-SegNet             & 83.66 & 1.821 & 2.937 & 76.14 & 1.353 & 1.783 & 79.39 & 1.647 & 2.342 \\ \hline
Pre-RecNet$_5$+Pre-SegNet        & 83.63 & 1.795 & 2.874 & 75.16 & 1.378 & 1.813 & 78.99 & 1.668 & 2.386 \\ \hline
SADFN$_5$-WOS+Pre-SegNet     & 83.85 & 1.782 & 2.838 & 75.84 & 1.357 & 1.762 & 79.25 & 1.661 & 2.364 \\ \hline
Liu\cite{27}+Pre-SegNet      & 84.08 & 1.776 & 2.814 & 76.30 & 1.335 & 1.724 & 79.37 & 1.661 & 2.357 \\ \hline
SADFN$_5$+Pre-SegNet         & \textbf{85.76} & \textbf{1.690} & \textbf{2.579} & \textbf{81.29} & \textbf{1.143} & \textbf{1.381} & \textbf{80.08} & \textbf{1.649} & \textbf{2.305} \\ \hline
Full-sampled+Pre-SegNet                & 87.30 & 1.596 & 2.328 & 86.89 & 0.973 & 1.092 & 80.76 & 1.617 & 2.225 \\ \hline
\end{tabular}
\end{table}

\section{Conclusion}

In this paper, we proposed a segmentation-aware deep fusion network (SADFN) for compressed sensing MRI. We showed the high-level supervision information can be effectively fused into deep neural network models to help the low-level MRI reconstruction. The multilayer feature aggregation is adopted to fuse cross-layer information in the MRI segmentation network and the feature fusion strategy is utilized to fuse cross-task information in the MRI reconstruction network. We prove the proposed SADFN architecture enables the reconstruction network aware of the contents it reconstructs and the function mapping can be significantly simplified. The SADFN model achieves state-of-the-art performance in CS-MRI and balance between accuracy and efficiency.

The SADFN model offers a novel approach to introduce supervision information from high-level vision tasks into low-level deep models. The proposed model can be easily extended to other computer vision tasks.


\bibliographystyle{splncs}
\bibliography{egbib}
\end{document}